\RequirePackage[T1]{fontenc}
\documentclass[letterpaper,10pt,conference]{ieeeconf}
\IEEEoverridecommandlockouts
\usepackage[font=small]{caption}
\usepackage{iftex}
\ifPDFTeX
    \usepackage[utf8]{inputenc}
\fi
\usepackage{amsmath,amssymb}
\usepackage{graphicx}
\usepackage{cite}
\usepackage{subcaption}
\usepackage{hyperref}
\usepackage{booktabs}
\usepackage{multirow}
\usepackage{tabularx}
\usepackage{xcolor}

\title{\LARGE\bf Communication-Aware Heterogeneous Graph Learning for Decentralized Multi-Human Multi-Robot Task Allocation}

\author{Ziqin Yuan$^{1}\dag$, Ruiqi Wang$^{1}\dag$, Baijian Yang$^{1}$, and Byung-Cheol Min$^{2}$
\thanks{$\dag$ Equal contribution}
\thanks{$^{1}$Purdue University, West Lafayette, IN, USA.}
\thanks{$^{2}$Indiana University Bloomington, Bloomington, IN, USA.}
\thanks{Correspondence: \texttt{wang5357@purdue.edu}; \texttt{minb@iu.edu}.}
}

\begin{document}
\maketitle

\begin{abstract}
Multi-human multi-robot (MH-MR) teams combine robotic autonomy with human expertise, but effective task allocation requires coordinating scarce, dynamically available human support with distributed robot execution. Limited robot-robot and human-robot communication further complicates this coupling by delaying information exchange and supervisory intervention. We introduce CommHG, a communication-aware heterogeneous graph learning framework for decentralized MH-MR task allocation. CommHG represents coupled human-robot-task interactions through local graphs conditioned on information availability and age. Learned communication actions enable robots to decide when and which operator to query, and when to share information with peers. Allocation and communication are jointly optimized through cooperative multi-agent reinforcement learning, allowing the team to acquire useful information while managing limited communication and supervisory resources. We also introduce a benchmark integrating heterogeneous humans and robots, dynamic tasks and operational states, and constrained robot-robot and bidirectional human-robot links. Experiments across heterogeneous teams with up to 16 robots, 6 humans, and 112 tasks show that CommHG improves timely weighted mission completion as coordination scale increases, outperforming the strongest baseline in the Large scenario. Website at: \url{https://sites.google.com/view/commhg/home}.
\end{abstract}

\section{Introduction}
\label{sec:introduction}

Multi-human multi-robot (MH-MR) teams are emerging as a promising approach to large-scale missions that combine robotic autonomy with human expertise. By coordinating multiple operators and heterogeneous robots, these teams can extend autonomous execution with selective human supervision and intervention~\cite{wang2023ita,wang2024ita,hoque2023fleet}. Realizing this potential requires effective allocation of both mission tasks and limited human support.
\begin{figure}[t]
    \centering
    \begin{subfigure}[b]{0.95\columnwidth}
        \centering
        \includegraphics[width=\linewidth]{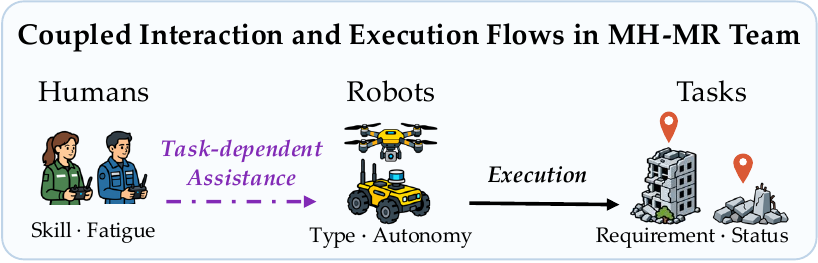}
        \caption{Human–robot–task interaction flow.}
        \label{fig:1a}
    \end{subfigure}
    \vspace{2mm}
    \begin{subfigure}[b]{0.95\columnwidth}
        \centering
        \includegraphics[width=\linewidth]{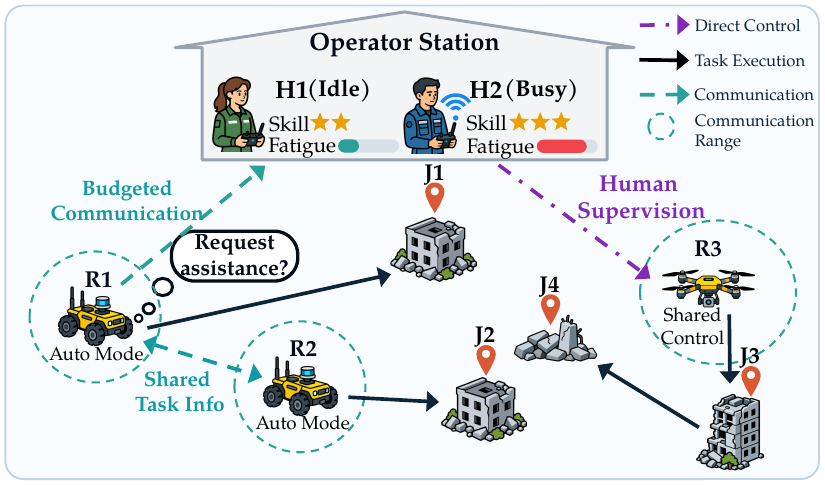}
        \caption{MH-MR collaboration under limited communication.}
        \label{fig:1b}
    \end{subfigure}
    \vspace{-5pt}
    \caption{Conceptual illustration of challenges in MH-MR coordination.}
    \label{fig:1}
    \vspace{-18pt}
\end{figure}

Compared with heterogeneous multi-robot allocation~\cite{peng2023gatar,ratnabala2025magnnet}, MH-MR allocation remains less explored. Transferring robot allocation methods requires explicitly accounting for the supervisory role of humans. In our setting, robots execute tasks, while humans are scarce resources whose intervention changes robot performance. As illustrated in Fig.~\ref{fig:1a}, this creates a coupled \emph{human-robot-task interaction flow}: a task determines the need for assistance, an operator's support changes its execution time and reliability, and allocating that operator reduces the support available to other robots. Human proficiency, evolving fatigue, and ongoing supervisory commitments further change the value and availability of assistance. Thus, treating a human simply as another independent task executor misses these dependencies.

Dedicated MH-MR methods have begun to address this coupling through heterogeneous initial allocation, dynamic reallocation, and load management under partial observability~\cite{wang2023ita,wang2024ita,yuan2025ata,wu2022load}. However, modeling changing states leaves open how to obtain useful updates when communication is limited. Centralized state reconstruction~\cite{yuan2025ata} and local belief updates~\cite{wu2022load} support decisions from available information, but do not by themselves select when and whom to query. For robots operating in the field (Fig.~\ref{fig:1b}), restricted robot-robot and human-robot communication can leave task progress, assignment intentions, and operator availability outdated or unknown. An allocation based on an old report may direct a robot toward a completed task or rely on an operator who is already occupied. Communication therefore needs to be part of the allocation decision. Although communication-constrained human assistance and learned communication scheduling have been studied~\cite{zhang2025flykites,ding2020i2c,niu2021magic}, their integration with coupled MH-MR allocation motivates our approach.

We propose \textbf{CommHG}, a \emph{communication-aware heterogeneous graph learning} framework for decentralized MH-MR task allocation. CommHG represents human supervisory state, robot execution state, and task demand through typed nodes and interaction-specific features. It addresses limited communication in two complementary ways. First, a \emph{communication-conditioned local graph} encodes only permitted observations and delivered records, retaining information age and knownness so that missing and stale information are represented during policy learning. Second, \emph{learned communication actions} allow robots to actively acquire and share information: an information-query head selects whether and which operator to query, while a broadcast gate selects when to communicate with robot neighbors. These decisions are trained jointly with task and supervisory allocation using cooperative multi-agent reinforcement learning. Queries and replies consume limited transmission resources and influence subsequent decisions only after delivery, allowing their value to be learned through downstream team performance.
The key distinction is that CommHG treats information acquisition as part of MH-MR allocation: it learns when an update is worth the limited communication resources it shares with human supervision.
Our contributions are:
\begin{itemize}
    \item Communication-conditioned heterogeneous graph representations. We represent coupled human-robot-task interactions while preserving receiver-specific information availability and age, enabling allocation policies to account for missing and outdated information.

    \item Joint learning of allocation and active communication. We jointly learn robot task allocation, human supervisory allocation, and communication under a shared team objective. Robots actively decide when and which operator to query, and when to broadcast to peers, to support subsequent allocation decisions.

    \item A benchmark for dynamic MH-MR coordination under limited communication. We integrate human and robot heterogeneity, evolving states and tasks, and constrained communication. Experiments demonstrate improved team performance, particularly in large-scale scenarios and under tighter communication constraints.
\end{itemize}

\section{Background and Related Work}
\label{sec:related_work}

\subsection{Task Allocation in MH-MR Teams}

Task allocation assigns work according to agent capabilities, task requirements, and resource availability. Heterogeneous multi-robot methods, including \cite{peng2023gatar} and \cite{ratnabala2025magnnet}, use graph-based policies to coordinate robots with different embodiments. Supervisory MH-MR teams introduce an additional dependency: allocating a human changes how a robot executes its task. Human proficiency determines which interventions are effective, while workload and fatigue change the support available over time. Effective allocation must therefore coordinate both robot-task assignments and access to human supervision.

Attention-based MH-MR methods model differences among humans, robots, and tasks for initial allocation\cite{wang2023ita,wang2024ita}. \cite{yuan2025ata} extends this direction to centralized conditional reallocation by reconstructing noisy or delayed states. \cite{wu2022load} studies decentralized load management under partial observability, while \cite{hoque2023fleet} allocates limited human supervision to improve robot policies through interactive fleet learning. These studies highlight human resources and changing team conditions. CommHG focuses on decentralized mission allocation in which acquiring the information needed to coordinate those resources is itself a learned decision. Its robot and human policies jointly select task and supervisory assignments using locally available information.

\subsection{Graph Representations for Team Coordination}

Graph neural networks represent entities as nodes and their relationships as edges, then aggregate neighboring features into decision-relevant embeddings~\cite{velickovic2018gat}, and have been applied to multi-agent coordination~\cite{jiang2020dgn}. Shared encoders and aggregation support varying team sizes, while heterogeneous graphs distinguish entity and relation types. HetNet uses type-aware graph processing for communication among heterogeneous agents~\cite{seraj2022hetnet}. Allocation graphs can also include tasks and resources: ScheduleNet represents scheduling relationships among tasks, robots, and spatial resources~\cite{wang2020schedulenet}, and CapAM learns task-graph embeddings conditioned on robot and mission information~\cite{paul2022capam}. These approaches underpin structured allocation representations.

CommHG adapts this relational representation to supervisory MH-MR coordination through task-to-robot, cached-peer-to-robot, and delivered-robot-report-to-human relations. Its graph is receiver-local: remote features come from received records, and relation features preserve task beliefs, supervisory context, and information age. The distinction between an interaction and its available evidence is important under limited communication. A robot may remain assigned to a task even when another agent has no current report of its progress. CommHG therefore conditions graph inputs on record availability and freshness, allowing policies to account for missing or stale information while learning coupled allocation decisions. Computational graph aggregation operates on stored records, and physical message delivery determines when those records change.

\begin{figure*}[t]
    \centering
    \includegraphics[width=\textwidth,pagebox=cropbox]{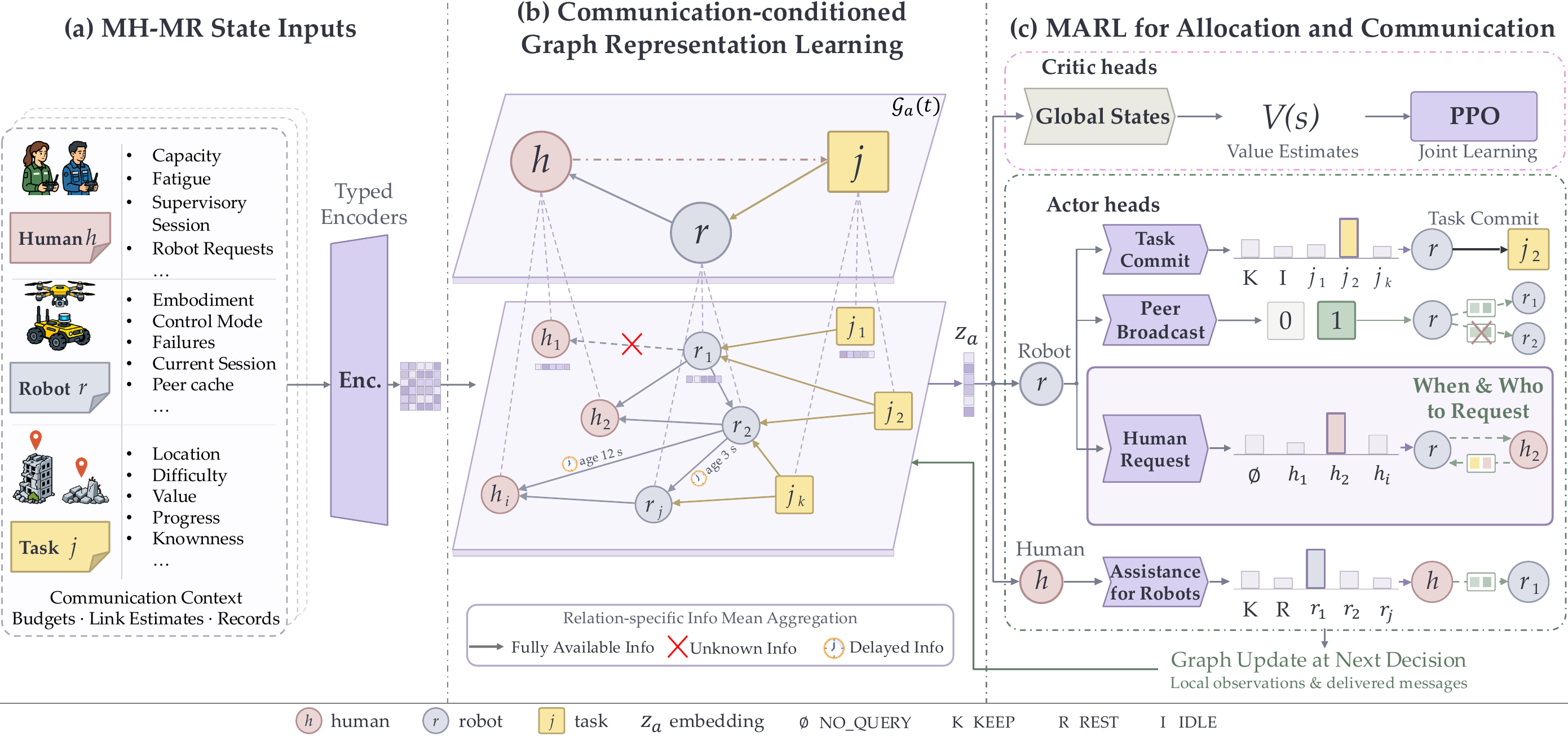}
\caption{\small Overview of CommHG for decentralized MH-MR task allocation under limited communication.
(a) Each decision maker uses its own state, public task attributes, received records, and local communication resources.
(b) Typed encoders and relation-specific aggregation represent human supervision, robot execution, and task demand. Local graphs retain information availability and age.
(c) Role-specific policies jointly learn task allocation, peer broadcasting, operator queries, and human supervisory allocation. Robots decide when and which human to query; humans decide which robot to assist. A centralized critic uses global state during PPO training, while execution relies on local observations. Actions affect the graph at subsequent decisions through locally observed changes and delivered messages.}
    \label{fig:CommHG_framework_rev}
    \vspace{-15pt}
\end{figure*}

\subsection{Active Communication and Information Queries}

Under partial observability, communication can improve coordination by supplying information unavailable to an agent's own sensors~\cite{foerster2016dial}. Active communication makes information access a policy choice, balancing its expected decision benefit against limited transmission resources. IC3Net learns communication gates~\cite{singh2019ic3net}, and SchedNet learns which agents access a shared communication medium~\cite{kim2019schednet}. MAGIC learns communication timing and partners together with graph-based message processing~\cite{niu2021magic}; T2MAC combines selective engagement with evidence-based message integration~\cite{sun2024t2mac}; TarMAC learns targeted messaging via attention~\cite{das2019tarmac}. I2C learns communication partners from their inferred influence on joint action values and uses a request-reply mechanism~\cite{ding2020i2c}. Thus, downstream decision value motivates selective communication in multi-agent learning.

CommHG integrates these choices into coupled supervisory allocation. Robots select whether and whom to query using local interaction embeddings and cached operator-resource reports, and a separate gate controls peer broadcasts. Queries compete with robot telemetry, and replies compete with human control commands, linking information acquisition to intervention resources. Because operators share station-side task evidence, recipient choice concerns response availability and control demand. A delivered reply updates information for reallocation, while human assistance requires a separate accepted control command. This differs from FlyKites' planning of relay topology, operator movement, and robot relay roles to restore connectivity for assistance~\cite{zhang2025flykites}. CommHG learns allocation and communication together under budgeted robot-robot and human-robot transport.

\subsection{Benchmarks for Dynamic Team Coordination}

Coordination benchmarks emphasize different aspects of team operation. FireCommander provides heterogeneous perception and action roles in a probabilistic environment~\cite{seraj2020firecommander}. Fleet-DAgger evaluates the allocation of human supervision for fleet learning~\cite{hoque2023fleet}, while ATA-HRL evaluates heterogeneous MH-MR allocation under dynamic and uncertain operational states~\cite{yuan2025ata}. These settings motivate evaluating human resource use alongside robot task performance.

Our benchmark combines human proficiency and fatigue, aerial and ground robots, changing tasks and failures, and budgeted robot-robot and bidirectional human-robot communication. The central challenge is their interaction: changing workloads alter the value of supervision while delayed information can prevent timely reassignment. Controlled variations in team scale, dynamics, and communication constraints support comparisons of graph representations and active communication under matched information access and resource budgets. Mission value and communication expenditure are reported separately to assess whether improved coordination justifies its information cost.

\section{Methodology}
\label{sec:methodology_rev}

CommHG learns task allocation and selective communication from receiver-local information. As shown in Fig.~\ref{fig:CommHG_framework_rev}, it encodes human, robot, and task observations into a communication-conditioned heterogeneous graph, then trains role-specific allocation and communication policies under a shared team objective.

\subsection{MH-MR State and Local Observations}
\label{subsec:CommHG_problem_rev}

We consider a cooperative partially observable decision process with asynchronous, variable-duration decisions, giving a semi-Markov formulation. Humans $h\in\mathcal H$ allocate supervision, robots $r\in\mathcal R$ execute tasks, and $j\in\mathcal T_t$ indexes tasks released by time $t$. The entity states comprise
\begin{equation}
\begin{aligned}
 \mathbf s_h^H &= [\mathbf p_h,c_h,f_h,u_h],\\
 \mathbf s_r^R &= [e_r,\mathbf p_r,b_r,m_r,x_r],\\
 \mathbf s_j^T &= [\mathbf p_j,k_j,D_j,v_j,d_j,\sigma_j,\eta_j].
\end{aligned}
\label{eq:CommHG_entity_states}
\end{equation}
Human features describe proficiency $\mathbf p_h$, cognitive capacity $c_h$, fatigue $f_h$, and control session $u_h$. Robot features describe embodiment $e_r$, pose $\mathbf p_r$, activity $b_r$, execution mode $m_r$, and assignment $x_r$. Task features describe location $\mathbf p_j$, type $k_j$, difficulty $D_j$, value $v_j$, deadline $d_j$, status $\sigma_j$, and progress $\eta_j$. Human proficiency depends on robot embodiment, and fatigue evolves with workload and rest. Human control changes robot travel, service performance, and failure risk; the robot's task determines the resulting supervisory workload. Allocation must therefore account for the coupled human-robot-task relationship and competition for limited operators.

We also consider the limited information flows across the team. Released task locations and announced attributes are public. Each actor observes its own state and communication resources. Robots additionally sense nearby tasks and their own assignments. Remote operational states and current task type, status, and progress are available only through permitted sensing or delivered records. A robot thus maintains task beliefs and cached peer and operator reports, each with age and knownness. Humans use delivered robot telemetry and a station-side shared task-evidence table. This table contains reports and release priors rather than a live global task map. Unreported remote changes remain hidden from the actor. The full simulator state, including physical states, assignments, in-transit messages, records, budgets, and leases, is available only to the centralized critic during training.

\subsection{Communication-Conditioned Graph Learning}
\label{subsec:CommHG_graph_rev}

The graph encoder learns a policy input that represents execution relationships together with the availability and age of their supporting information. For each actor $a$, we construct
\begin{equation}
 \mathcal G_a(t)=\bigl(\mathcal V_a,\mathcal E_a,
                       \mathbf X_a,\mathbf F_a\bigr),
 \qquad
 \mathbf z_a=\operatorname{GNN}_{\theta,\operatorname{role}(a)}
                       \bigl(\mathcal G_a(t)\bigr).
\label{eq:CommHG_local_graph}
\end{equation}
Node features $\mathbf X_a$ encode the actor's own state, public task attributes, and available remote records. Edge features $\mathbf F_a$ describe pairwise relationships: task-robot edges carry distance, compatibility beliefs, assignment context, believed task status/progress, and evidence age; peer-robot edges carry cached relative geometry and record age; robot-report-human edges carry reported task context and local supervisory-session information. Received task tips enter both these relationship features and task scoring.

Three directed computational relations govern aggregation: task-to-robot, cached-peer-to-robot, and delivered-robot-report-to-human. Typed encoders produce 128-dimensional node embeddings. Relation-specific MLPs produce 64-dimensional messages, which are mean-pooled separately by relation. Two parameter-shared residual updates with role-specific MLPs and layer normalization yield $\mathbf z_a$, following graph-based multi-agent representation learning~\cite{nayak2023informarl}. Both rounds operate on the receiver's existing records and require no additional transmission.

The representation is communication-conditioned because delivery determines which remote records enter aggregation and how current their contents are. Unknown peer/report records are masked; missing task fields retain explicit unknown indicators while public attributes remain available. Delayed records retain their original evidence times, and relaying does not refresh them. Newly delivered, supported evidence updates a task field only when newer than its stored evidence; completion is terminal. These features allow the policy to distinguish missing, stale, and recently observed information. Local observations and delivered messages update the graph for subsequent decisions, while explicit records provide temporal memory without a recurrent actor.

\subsection{Cooperative MARL for Allocation and Communication}
\label{subsec:CommHG_policy_rev}

CommHG uses centralized training with decentralized execution (CTDE)~\cite{lowe2017maddpg} and cooperative PPO~\cite{yu2022mappo,schulman2017ppo}. Humans share one actor and robots another. Their action spaces are
\begin{equation}
\begin{aligned}
 a_r &= (a_r^T,g_r,a_r^{RH}),\qquad a_r^{RH}=(q_r,\alpha_r),\\
 a_r^T &\in\{\mathrm{KEEP},\mathrm{IDLE}\}\cup\mathcal T_t,\\
g_r,\alpha_r &\in\{0,1\},\qquad q_r\in\{\varnothing\}\cup\mathcal H,\\
 a_h &\in\{\mathrm{KEEP},\mathrm{REST}\}\cup\mathcal R,
\end{aligned}
\label{eq:CommHG_actions}
\end{equation}
where $g_r$ is the robot-robot (RR) broadcast gate, $q_r$ is the robot-human (RH) query recipient, and $\alpha_r$ is the conditional assistance-request action; $\varnothing$ denotes no query. Robot decisions comprise three functional categories. \emph{Task allocation} scores each task from the robot embedding, task encoding, and pairwise features. \emph{RR communication} uses a binary gate to broadcast to transport-selected neighbors. \emph{RH interaction} selects either an information query to a human or an advisory assistance request. The query head scores recipients using their last-received resource reports, including tokens, queue/occupancy state, age, and knownness. Idle robots may query. The binary assistance-request head is sampled only when no query is selected and a held task, lease/bid status, and uplink resources permit it. Humans independently score robots using delivered telemetry and supervisory context, then select control, continuation, or rest, while an assistance request is not a prerequisite for control.

Communication has a physical opportunity cost. RR links have limited range and out-degree, terrain-dependent loss, and delay. RH traffic uses separate per-robot uplink and per-human downlink budgets with delay, loss, and expiry. Queries and assistance requests displace ordinary telemetry opportunities, while replies compete with higher-priority control commands. A query returns the selected human's resource report and at most one task tip from shared station evidence. Recipient selection therefore concerns response availability and competing control demand. Replies may be stale or empty and affect allocation only after delivery; queries do not reserve human assistance. Control takes effect after command acceptance and handover, and expires with its lease. Joint learning can consequently favor communication that improves subsequent allocation enough to justify the resources it consumes.

\subsection{Shared Team Reward}
\label{subsec:CommHG_learning_rev}

All actors maximize the same discounted mission return. The reward for a one-second transition is
\begin{equation}
\begin{aligned}
 r_t={}&
 \underbrace{\sum_{j\in\mathcal C_t}v_j\ell_j(t+1)
 -\lambda_P\frac{\sum_{j\in\mathcal P_t}v_j}{V_0}}_{r_t^M:\ \text{mission}}\\
 &+\underbrace{\left(-C_t^{\rm switch}
 -C_t^{\rm incompat}-C_t^{\rm conflict}\right)}_{r_t^C:\ \text{coordination}}.
\end{aligned}
\label{eq:CommHG_reward_rev}
\end{equation}
Here $\mathcal C_t$ contains tasks completed during the transition, $v_j$ is task value, and $\ell_j(t)=\exp[-\max(0,t-d_j)/\tau_j]$ retains full value through deadline $d_j$ and discounts late completion with scale $\tau_j$. The pending set $\mathcal P_t$ contains released tasks still incomplete after the transition; $\lambda_P=0.01$ and $V_0$ is total mission task value, including later releases, used only for reward normalization. Coordination costs penalize non-exempt human switches, incompatible selections, and control conflicts. No byte penalty or query bonus is added. Communication is learned through its downstream team return under hard resource constraints.

\subsection{Training.}
CommHG follows centralized training with decentralized execution. During training, a separate graph critic observes the current states of all humans and robots, released tasks, assignments, and communication resources. It aggregates these features into a global context and combines this context with the critic's embedding of each evaluated agent. Four shared output heads estimate mission and coordination returns for the two roles: robot and human. These heads provide role-specific training signals for the same team objective. During execution, each actor uses only its permitted local observations and received records.

Because agents may make decisions at different times, we compute advantages over the physical time elapsed between consecutive decisions using time-aware GAE~\cite{schulman2016gae}. Mission and coordination advantages are added before normalization to guide all action heads toward the combined team return. For each robot decision, the likelihood ratios of the sampled action components are multiplied, and PPO clipping is applied once to this joint ratio. Human policies use the likelihood ratio of their supervisory action. Policy losses are first averaged within each role and then equally across roles, preventing the more numerous robots from dominating training. Components fixed by eligibility constraints contribute neither policy likelihood nor entropy.

Two shared mechanisms support feasible execution. A task resolver prevents multiple robots from owning the same task, while a one-time initializer uses public task information to distribute initial targets. During the bounded initialization phase, prescribed task choices are excluded from policy updates, but eligible communication actions and value estimates continue to be trained. Subsequent allocation decisions are made by the local actors under these initialization, shared-station, and task-arbitration assumptions.

\providecommand{\CommHGexpName}[2]{%
  \ifdefined\textcolor\textcolor[rgb]{#1}{\textbf{#2}}%
  \else\textbf{#2}\fi}
\providecommand{\CommHGexpGreedy}{\CommHGexpName{0.352941,0.368627,0.392157}{Greedy}}
\providecommand{\CommHGexpMAPPO}{\CommHGexpName{0.368627,0.474510,0.349020}{MAPPO}}
\providecommand{\CommHGexpIPPO}{\CommHGexpName{0.529412,0.411765,0.207843}{IPPO}}
\providecommand{\CommHGexpInforMARL}{\CommHGexpName{0.325490,0.458824,0.537255}{InforMARL}}
\providecommand{\CommHGexpATA}{\CommHGexpName{0.588235,0.368627,0.392157}{ATA-HRL}}
\providecommand{\CommHGexpCommHG}{\CommHGexpName{0.470588,0.376471,0.615686}{CommHG}}

\section{Experiment}
\label{sec:experiment}
\label{sec:experiments_results}

\subsection{MH-MR Simulation Benchmark}
\label{subsec:experiment_setup}

\begin{figure}[t]
    \centering
    \includegraphics[width=\columnwidth]{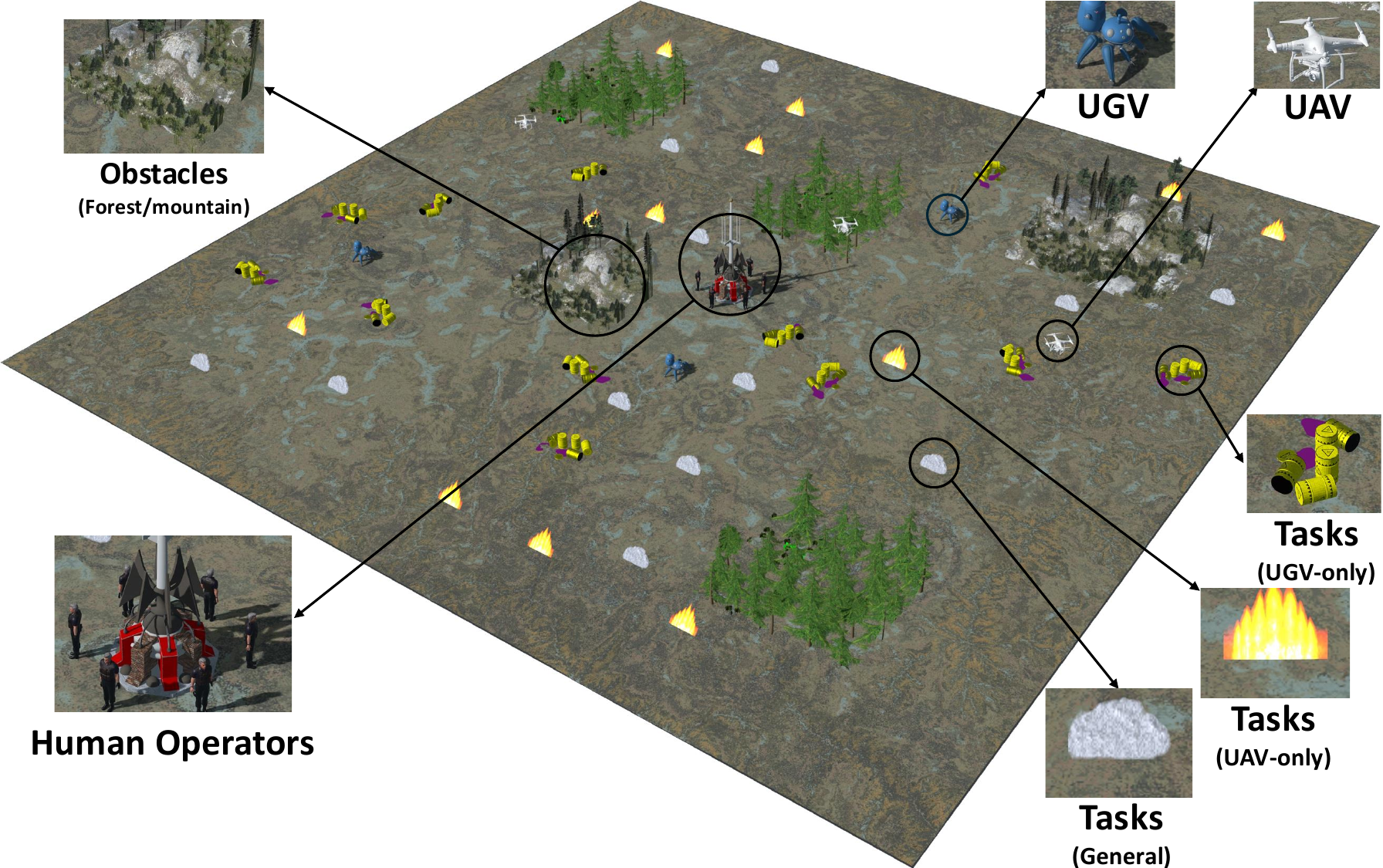}
    \caption{Overview of the MH-MR benchmark 3D simulation environment.}
    \label{fig:benchmark_overview}
    \vspace{-18pt}
\end{figure}

We introduce a simulation benchmark for communication-constrained MH-MR allocation, motivated by environmental surveillance missions. Aerial and ground robots service spatially distributed points of interest (POIs), while humans at a supervisory station selectively control robots. The benchmark couples two resource-allocation problems: assigning robots to mission tasks and assigning limited human capacity to those robots. The value of a robot-task pairing therefore depends not only on mobility and compatibility, but also on whether suitable human support can be obtained and whether the information used to request it is current. Figure~\ref{fig:benchmark_overview} shows a 3D rendering of the environment's components, namely the obstacles (forest and mountain), the station with its human operators, UAVs and UGVs, and the UAV-only, UGV-only, and general tasks.

\paragraph{Spatial world and mission scale}
The simulator operates on a planar map with spatially distributed POIs
and obstacle footprints. UGVs navigate around obstacles, whereas UAVs
use an abstract fly-over model. Task allocation is separated from
low-level navigation: a policy selects a task, and the simulator advances
travel and service. The 3D scene is a visualization of these states,
not a physics-based flight or manipulation simulation, and its rendered
images are not policy inputs. Table~\ref{tab:env_settings} lists Small,
Medium, and Large scenarios with 22, 56, and 112 tasks. Map extent,
team size, and mission duration increase together to test different
spatial and coordination scales.

\paragraph{Heterogeneous robots and humans}
Robots differ in travel speed and task compatibility. Both embodiments
can service general tasks, but UAV-only and UGV-only tasks require the
corresponding robot type. Humans are simulated supervisory decision
makers, not additional mobile task executors. Their embodiment-dependent
proficiencies comprise generalist, UAV-specialist, and UGV-specialist
profiles, so no human is necessarily the best choice for every robot.
Cognitive capacity affects fatigue accumulation and recovery, while fatigue
in turn changes the benefit of continued control. A human controls at
most one robot, and a robot receives control from at most one human.
Control can improve travel and service performance and reduce malfunction
risk, but handover and switching costs discourage instantaneous,
arbitrarily frequent reassignment. Robots can also execute autonomously,
making human support a scarce performance-enhancing resource rather
than a universal prerequisite for task completion.

\paragraph{Task execution and operational dynamics}
After reaching a compatible POI, a robot accumulates service progress
at a rate determined by task difficulty, operating mode, and effective
human support. A successful service attempt completes the task, a failed
attempt leaves it unfinished. This differs from a robot malfunction,
which removes that robot for the rest of the mission. Later task releases
introduce new work, and an unserviced general task can change to a
robot-exclusive type. Together with evolving human fatigue, these events
can invalidate an earlier allocation. Tasks carry values and soft
deadlines: late completion remains useful but earns discounted value.
The benchmark therefore rewards timely reallocation, not simply eventual
completion or short travel distance.

\paragraph{Communication and local information}
Robot-robot communication has limited range and out-degree, with delay,
loss, and terrain-dependent attenuation. Human-robot traffic additionally
uses separate robot uplink and human downlink budgets, and messages can
expire. Telemetry supplies humans with robot reports, while delivered
control commands establish or renew supervision. Information queries
share uplink resources with telemetry, and their replies compete with
control traffic. Thus, communicating more is not equivalent to obtaining
more immediately usable information. Released task locations and
announced attributes are public, but remote task progress, completion,
and changed compatibility must be learned through permitted observation
or delivered evidence. Human replies draw on reported information rather
than a perfect global task map. These restrictions distinguish acquiring
an update from obtaining assistance and make information age relevant
to allocation.

\begingroup
\begin{table}[t]
\centering
\caption{Simulation environment settings. Scenario-dependent settings are listed above the shared settings. Speed and proficiency pairs are ordered as (UAV, UGV). Proficiency is a service-time multiplier; lower values indicate faster operators on the corresponding robot type.}
\label{tab:env_settings}

\footnotesize
\setlength{\tabcolsep}{3pt}
\setlength{\aboverulesep}{1pt}
\setlength{\belowrulesep}{1pt}
\renewcommand{\arraystretch}{1.0}

\begin{tabularx}{\linewidth}{@{}Xccc@{}}
\toprule
\textbf{Setting} & \textbf{Small} & \textbf{Medium} & \textbf{Large} \\
\midrule
Map side (m) & 500 & 1000 & 2000 \\
Tasks & 22 & 56 & 112 \\
Time limit (s) & 600 & 1200 & 2400 \\
UAVs / UGVs / operators & 2 / 2 / 2 & 4 / 4 / 3 & 8 / 8 / 6 \\
\end{tabularx}
\par\nointerlineskip
\begin{tabularx}{\linewidth}{@{}Xl@{}}
\midrule
\textbf{Shared setting} & \textbf{Value} \\
\midrule
Autonomous speed (m\,s$^{-1}$) & $(18,\,5)$ \\
Controlled speed (m\,s$^{-1}$) & $(25,\,8)$ \\
Proficiency: generalist & $(0.90,\,0.90)$ \\
Proficiency: UAV specialist & $(0.65,\,1.20)$ \\
Proficiency: UGV specialist & $(1.20,\,0.65)$ \\
R--R range / out-degree / drop & $0.30$ map / $4$ / $0.05$ \\
Uplink budget / query size / TTL & 192\,B / 32\,B / 24\,s \\
\bottomrule
\end{tabularx}
\vspace{-15pt}
\end{table}
\endgroup

\subsection{Compared Methods}
\label{subsec:compared_methods}

We compare \CommHGexpCommHG{} with five baselines:
\begin{itemize}
    \setlength{\itemsep}{2pt}
    \setlength{\parsep}{0pt}
    \item \CommHGexpGreedy{} selects feasible assignments by immediate estimated utility.
    \item \CommHGexpMAPPO{}~\cite{yu2022mappo} uses centralized value learning with decentralized PPO actors.
    \item \CommHGexpIPPO{}~\cite{yu2022mappo} trains PPO policies and value functions independently using local information.
    \item \CommHGexpInforMARL{}~\cite{nayak2023informarl} uses graph-based neighborhood aggregation for multi-agent coordination.
    \item \CommHGexpATA{}~\cite{yuan2025ata} combines initial allocation, conditional reallocation, and state reconstruction for heterogeneous MH-MR teams.
\end{itemize}

We further evaluate three ablations of \CommHGexpCommHG{}:
\begin{itemize}
    \setlength{\itemsep}{2pt}
    \setlength{\parsep}{0pt}
    \item \textit{NoComm} replaces communication-conditioned graph connectivity with a fixed relation topology, retaining local information-access restrictions and physical communication.
    \item \textit{NoQuery} removes the human information-query head while retaining human supervision and other communication mechanisms.
    \item \textit{NoGraph} removes graph message passing while retaining allocation and communication action heads.
\end{itemize}

\subsection{Evaluation Metric and Protocol}
\label{subsec:evaluation_metric}

Our primary metric is the \emph{Timely Weighted Completion Ratio}
(TWCR): the fraction of the mission's total task value realized through
completed work, after discounting late completion. Let $\mathcal T$ be
the full mission task set, including later releases, and let
$\mathcal C\subseteq\mathcal T$ contain tasks successfully completed
by the end of the episode. For task $j$, $v_j>0$ is its base value,
$t_j^{\rm done}$ its completion time, $d_j$ its deadline, and
$\tau_j>0$ its late-value decay timescale. We calculate
\begin{equation}
    \mathrm{TWCR}
    =\frac{\sum_{j\in\mathcal C}v_j\ell_j(t_j^{\rm done})}
           {\sum_{j\in\mathcal T}v_j},
    \label{eq:twcr_evaluation}
\end{equation}
with the explicit deadline discount
\begin{equation}
    \ell_j(t)=
    \begin{cases}
        1, & t\leq d_j,\\
        \exp\!\left[-(t-d_j)/\tau_j\right], & t>d_j.
    \end{cases}
    \label{eq:twcr_deadline_discount}
\end{equation}
The denominator equals $V_0$ in the training objective and is fixed by
the mission, not by the tasks a policy chooses to attempt. On-time
completion contributes full value, late completion contributes discounted
value, and an incomplete task contributes zero while remaining in the
denominator. Partial service alone does not count as completion.
Consequently, $0\leq\mathrm{TWCR}\leq1$, with one attained when all
mission value is completed by its deadlines. TWCR is neither a raw task
completion fraction nor an episode-level binary success rate: it weights
both task importance and completion timeliness.

\begin{figure*}[t]
    \centering
    \includegraphics[width=\textwidth,pagebox=cropbox]{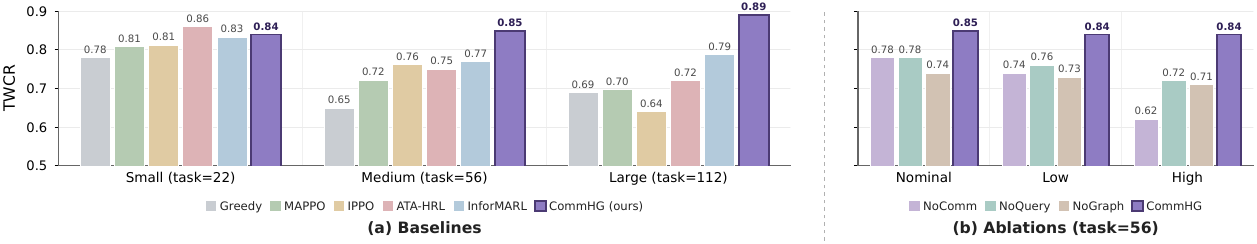}
    \vspace{-15pt}
    \caption{Mean TWCR (a) Baseline comparison across Small, Medium, and Large scenarios. (b) Ablations in the Medium scenario under increasing communication limitation. Numbers above bars indicate mean values.}
     \vspace{-15pt}
    \label{fig:main_results}
\end{figure*}

For every learned method, Figure~\ref{fig:main_results} reports means
over all 10 training seeds. The Greedy is a non-learning reference. Baseline comparisons cover all three mission scales, while the communication ablations keep the Medium team and task scale fixed. TWCR uses the simulator's recorded completions, not cached task beliefs or shaped training return: it excludes pending-work and coordination penalties and provides no direct credit for queries, transmitted bytes, or time spent under human control. Communication can therefore improve this metric only through better or timelier mission completion.

\section{Results and Analysis}
\label{sec:results_analysis}

\subsection{Main Results}
\label{subsec:main_results}

Figure~\ref{fig:main_results}(a) compares TWCR across three mission scales. \CommHGexpCommHG{} is competitive on Small, where \CommHGexpATA{} achieves the highest score, and outperforms all baselines on Medium and Large. Its advantage over the strongest competing baseline reaches approximately 10 percentage points on Large. This increasing advantage suggests that CommHG becomes particularly effective as more robots compete for tasks, human support, and communication resources.

\CommHGexpCommHG{} outperforms \CommHGexpGreedy{} across all three scales. Greedy allocation favors immediate estimated utility, which may overlook how assigning an operator or selecting a task affects subsequent team decisions. Such choices are particularly vulnerable to outdated information about task progress and human availability. CommHG instead learns allocation and communication through their long-term team consequences, allowing useful updates and future resource competition to influence current decisions.

Both \CommHGexpMAPPO{} and \CommHGexpIPPO{} also remain below \CommHGexpCommHG{}, particularly in larger missions. MAPPO provides centralized value information during training, but its actors still need to coordinate from limited observations at execution. CommHG supplies a structured representation of task, robot, and supervisory relationships together with learned communication actions. IPPO additionally relies on independent value learning, making it harder to account for the team-wide consequences of competing local choices. CommHG combines centralized training signals with relational actor inputs and selective information exchange to address these coupled decisions.

\CommHGexpInforMARL{} is the closest competitor on Medium and Large, indicating that graph-based information aggregation provides a strong foundation for coordination. \CommHGexpCommHG{} further conditions graph connectivity on receiver-specific record availability and retains information age, helping its policies distinguish usable evidence from missing or outdated reports. Its active queries also allow robots to obtain updates relevant to reallocation. These mechanisms offer a plausible explanation for the remaining performance gap, which the following ablations examine more directly.

\CommHGexpATA{} achieves the best result on Small, showing the benefit of specialized MH-MR allocation with dynamic state modeling. However, its centralized state reconstruction addresses uncertainty in the information supplied to the allocator without explicitly learning when and whom to query. \CommHGexpCommHG{} couples local allocation with control over information acquisition, which may become more valuable as teams scale and supervisory commitments must be tracked through limited communication. Its stronger performance on Medium and Large is consistent with this advantage.

\subsection{Ablation Study}
\label{subsec:communication_ablations}

Figure~\ref{fig:main_results}(b) evaluates the three ablations defined in Sec.~\ref{subsec:compared_methods} on the Medium scenario. Nominal, Low, and High communication limitation progressively reduce communication range and increase the combined probability of unavailable or delayed communication from 10\% to 20\% to 30\%, with equal shares assigned to the two mutually exclusive outcomes. Team and task scale remain fixed.

\paragraph{Communication-conditioned graph connectivity.}
CommHG maintains nearly unchanged TWCR as communication becomes more restrictive, while NoComm exhibits the largest deterioration. Their gap reaches approximately 22 percentage points under High limitation. This result supports adapting graph connectivity to receiver-specific record availability: as communication becomes less dependable, restricting aggregation to available sources helps align the learned representation with evidence accessible to each actor.

\paragraph{Active information acquisition.}
NoQuery performs below CommHG, and its disadvantage grows under tighter communication constraints. Since human supervision remains available, this comparison supports the value of actively requesting information in addition to receiving assistance. Queries can reveal changes in operator availability or task state that would otherwise remain unknown, enabling robots to revise allocations using updated evidence. These benefits arise through subsequent team performance, without a query-specific reward.

\paragraph{Relational representation.}
NoGraph performs worse than CommHG even under Nominal limitation, indicating that allocation and communication heads benefit from relational aggregation. The graph connects robot capabilities, task demands, and supervisory context, providing structured inputs for their coupled decisions. NoGraph's smaller additional decline under tighter constraints accompanies lower overall TWCR and therefore does not imply better coordination.

Together, the ablations support the complementary roles of relational representation, communication-conditioned connectivity, and active querying in maintaining team performance under limited communication.

\subsection{Case Study}
\label{subsec:query_case_study}
\begin{figure}[h]
    \centering
    \includegraphics[width=\columnwidth]{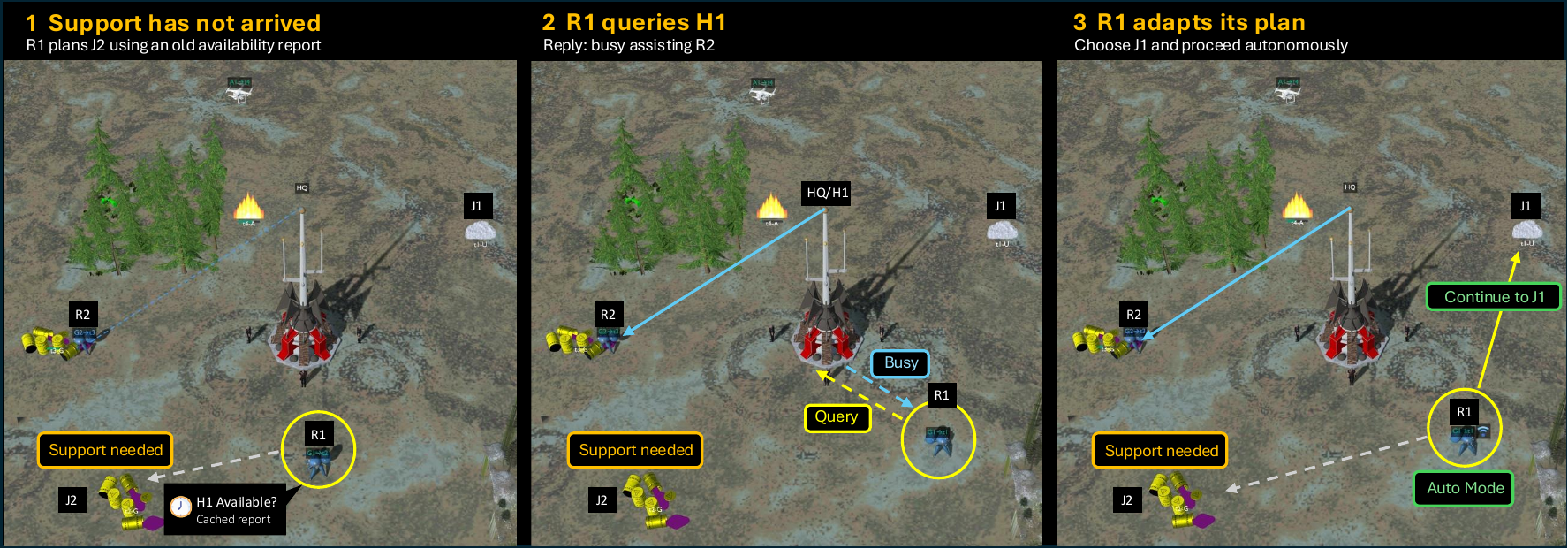}
    \caption{Active querying in an execution sequence. R1 expects H1's support for J2 (left), learns that H1 is assisting R2 (middle), and selects J1 for autonomous execution (right). Delivered information changes allocation without reallocating human support.}
     \vspace{-15pt}
    \label{fig:query_case_study}
\end{figure}

Figure~\ref{fig:query_case_study} visualizes a case illustrating how active querying supports timely reallocation. R1 initially plans to execute the difficult task J2 with H1's support, based on a cached availability report. When the expected support does not arrive, R1 queries H1 and learns that the operator is busy assisting R2. Upon receiving the reply, R1 updates its local information and selects J1, which it can execute autonomously, while H1 continues supporting R2.

This case highlights how \CommHGexpCommHG{} uses active queries to resolve uncertainty about task-relevant human resources. By querying the operator on whom its plan depends, R1 obtains information that enables it to revise its task choice and continue productive execution without disrupting ongoing human support.

\section{Conclusion}
\label{sec:conclusion}

We presented CommHG, a framework for decentralized MH-MR task allocation that jointly learns human supervision, robot task execution, and selective communication. Communication-conditioned heterogeneous graphs represent coupled interactions using locally available information, while active queries enable robots to obtain updates relevant to subsequent allocation decisions. Experiments on our benchmark demonstrate improved team performance, particularly at larger scales and under tighter communication constraints. These findings highlight the value of coordinating information acquisition with the allocation of limited human and robotic resources.

\bibliographystyle{IEEEtran}
\bibliography{main}
\end{document}